\documentclass[runningheads]{llncs}
\usepackage[T1]{fontenc}
\usepackage{graphicx}
\usepackage{booktabs} 
\usepackage{array}
\newcolumntype{P}[1]{>{\raggedright\arraybackslash}p{#1}}

\usepackage{caption}
\usepackage{subcaption}
\usepackage{orcidlink}
\usepackage{adjustbox}
\hypersetup{
    colorlinks=true,
    linkcolor=blue,
    urlcolor=blue,
    citecolor=black,
    filecolor=black,
    linkbordercolor=white, 
    urlbordercolor=white,
    citebordercolor=white,
    filebordercolor=white,
}

\begin{document}
%

\title{Towards Interactive Lesion Segmentation in Whole-Body PET/CT with Promptable Models}

%
%
\author{Maximilian Rokuss\inst{1,2,3}\orcidlink{0009-0004-4560-0760} \and 
        Yannick Kirchhoff\inst{1,2,3}\orcidlink{0000-0001-8124-8435} \and
        Fabian Isensee\inst{1,4}\orcidlink{0000-0002-3519-5886} \and
        Klaus H. Maier-Hein\inst{1,4,5}\orcidlink{0000-0002-6626-2463}
}
\authorrunning{M. Rokuss et al.}
\titlerunning{Interactive Lesion Segmentation in PET/CT}

\institute{
German Cancer Research Center (DKFZ) Heidelberg,\\Division of Medical Image Computing, Heidelberg, Germany
\and
HIDSS4Health - Helmholtz Information and Data Science School for Health, Karlsruhe/Heidelberg, Germany\and
Faculty of Mathematics and Computer Science,\\Heidelberg University, Heidelberg, Germany\and
Helmholtz Imaging, DKFZ, Heidelberg, Germany\and
Pattern Analysis and Learning Group, Department of Radiation Oncology, Heidelberg University Hospital, Heidelberg, Germany
\email{maximilian.rokuss@dkfz-heidelberg.de}
}
\maketitle              
\begin{abstract}
Whole-body PET/CT is a cornerstone of oncological imaging, yet accurate lesion segmentation remains challenging due to tracer heterogeneity, physiological uptake, and multi-center variability. While fully automated methods have advanced substantially, clinical practice benefits from approaches that keep humans in the loop to efficiently refine predicted masks. The autoPET/CT IV challenge addresses this need by introducing interactive segmentation tasks based on simulated user prompts. In this work, we present our submission to Task 1. Building on the winning autoPET III nnU-Net pipeline, we extend the framework with promptable capabilities by encoding user-provided foreground and background clicks as additional input channels. We systematically investigate representations for spatial prompts and demonstrate that Euclidean Distance Transform (EDT) encodings consistently outperform Gaussian kernels. Furthermore, we propose online simulation of user interactions and a custom point sampling strategy to improve robustness under realistic prompting conditions. Our ensemble of EDT-based models, trained with and without external data, achieves the strongest cross-validation performance, reducing both false positives and false negatives compared to baseline models. These results highlight the potential of promptable models to enable efficient, user-guided segmentation workflows in multi-tracer, multi-center PET/CT. Code is publicly available at \url{https://github.com/MIC-DKFZ/autoPET-interactive}.

\end{abstract}
\section{Introduction}

Positron Emission Tomography combined with Computed Tomography (PET/CT) is an essential modality in oncological imaging, providing both metabolic and anatomical information for tumor detection, staging, and therapy monitoring. Accurate lesion segmentation is central to these tasks but remains highly time-consuming and prone to inter-observer variability when performed manually. Automated segmentation offers substantial potential to accelerate radiological workflows and enable comprehensive quantification. However, its effectiveness is limited by physiological variability, tracer-dependent uptake patterns, and heterogeneity across imaging centers. These factors complicate the distinction between physiological and malignant uptake, particularly in a multi-tracer setting. \\

\noindent To advance robust automated methods, the autoPET challenge series has progressively expanded its scope. The \href{https://autopet-iii.grand-challenge.org}{autoPET III challenge} introduced large-scale multi-tracer, multi-center datasets, combining 1014 FDG PET/CT scans \cite{gatidis2020fdgpetct} with 597 PSMA tracer scans \cite{jeblick2024psmapetct}, and established the current benchmark for fully automated lesion segmentation. The follow-up \href{https://autopet-iv.grand-challenge.org}{autoPET/CT IV challenge} shifts the focus toward interactive, human-in-the-loop segmentation. In \textit{Task 1}, algorithms are evaluated under increasing levels of simulated user input in the form of foreground and background clicks, ranging from fully automated predictions to guidance with up to 10 lesion and 10 background clicks per scan. This setup enables a systematic investigation of how interactive conditioning improves segmentation quality in whole-body PET/CT, while avoiding the need for exhaustive manual annotation of every lesion. \\

\noindent This focus reflects a broader paradigm shift in computer vision and medical image analysis toward promptable models. Initiated by the Segment Anything Model (SAM)~\cite{kirillov2023segany}, interactive prompting has rapidly been adapted for medical imaging tasks~\cite{nninteractive}, including lesion segmentation~\cite{LesionLocator}, where sparse user inputs provide informed cues to resolve ambiguities in complex anatomical contexts. AutoPET/CT IV provides an opportunity to benchmark such approaches in the setting of multi-tracer, multi-center PET/CT. \\

\noindent In this work, we present our submission to Task~1 under the team name \textit{LesionLocator}, inspired by the LesionLocator model for universal promptable lesion segmentation~\cite{LesionLocator}. Building on the nnU-Net framework \cite{isensee2021nnu} and the optimized winning solution from autoPETIII~\cite{rokuss2024fdg}, we extend the automatic segmentation pipeline to incorporate simulated user interactions. Specifically, the model still provides auto-segmentation masks if no clicks are provided but leverages the user prompts to refine and improve the initial results.

\section{Methods}

Our approach builds directly on the winning solution from the previous autoPET III challenge~\cite{rokuss2024fdg}, which is based on the nnU-Net framework \cite{isensee2021nnu} extended with several key advancements: (i) the more powerful ResEncL U-Net architecture preset \cite{isensee2024nnu}, (ii) a large-scale multi-modal pretraining and fine-tuning strategy that enabled robust cross-domain feature learning, (iii) organ supervision to mitigate false positives in physiologically active regions, and (iv) tailored data augmentation including PET–CT misalignment simulation. Together, these components substantially improved generalization across tracers, centers, and patient populations. The model uses the \verb+3d_fullres+ configuration, resamples all images to a common spacing of \verb+[3, 2.04, 2.04]+ and normalizes both modalities with the default CT normalization scheme. We train with a batch size of 2 for 1000 epochs and a uniform patch size of \verb+192x192x192+. We use an initial learning rate of \verb+1e-3+ for fine-tuning where we also experiment with MultiTalent~\cite{ulrich2023multitalent} pretraining on more datasets. Additional details can be found in Table \ref{tab1}.\\

\noindent We retain this strong foundation and introduce modifications to address the novel human-in-the-loop setting. Specifically, we adapt the model to leverage interactive click-based inputs by adding two additional input channels encoding foreground (positive) and background (negative) prompts, while keeping the underlying architecture, loss, and hyperparameters unchanged. The key challenges lie in (i) how to represent clicks as model inputs and (ii) how to realistically simulate user interactions during training.\\

\noindent \textbf{Click representation.} A central design choice in promptable segmentation models is how spatial prompts are encoded for the network. In our approach, prompts are provided as additional input channels to the U-Net, but the critical question is how to represent a point at a given spatial location. We experimented with Gaussian kernels of varying standard deviations $\sigma$, normalized to unit volume, as well as Euclidean Distance Transforms (EDT) with different scaling factors inspired by nnInteractive \cite{nninteractive}. The EDT-based representation consistently outperformed Gaussian kernels, likely because the normalization of the latter results in low-intensity voxel values that are not effectively captured by the network.\\

\noindent \textbf{Simulating user interaction.} To improve variability and generalization, we generate prompts online during training rather than relying solely on precomputed point locations. Prompts are simulated after data augmentation and directly within the sampled training patches, ensuring that their spatial representation remains consistent while exposing the network to diverse imaging conditions. For positive and negative clicks respectively, we simulate between zero and ten prompts per patch, sampled with a logarithmic probability that favors fewer points, i.e. up to twenty total prompts are possible but rarely occur. For validation, we restrict ourselves to the official precomputed prompts to guarantee comparability across models. Prompt locations are primarily simulated using the official challenge code to match the distribution expected at test time, but we extend this with an additional implementation to introduce further variability, using an 80\%/20\% sampling strategy between the two. The main difference is that our simulated prompts are less restricted to centers or borders of lesions and can instead appear anywhere inside the target region, with a higher probability of being sampled near the lesion core.

\begin{figure}[t]
    \centering
    \begin{subfigure}[t]{0.32\textwidth}
        \centering\includegraphics[width=\textwidth]{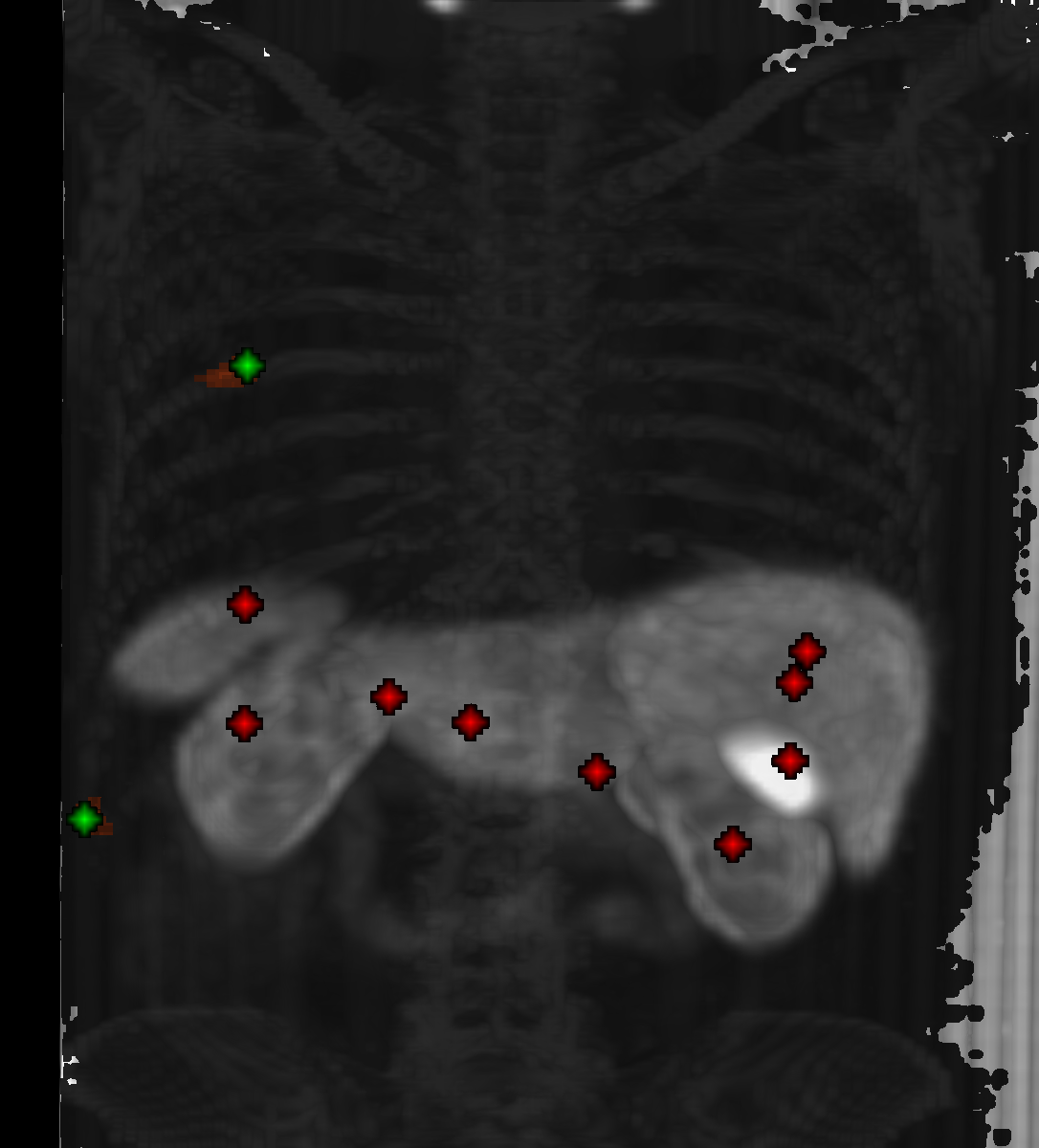}
    \end{subfigure}
    \begin{subfigure}[t]{0.32\textwidth}
        \centering\includegraphics[width=\textwidth]{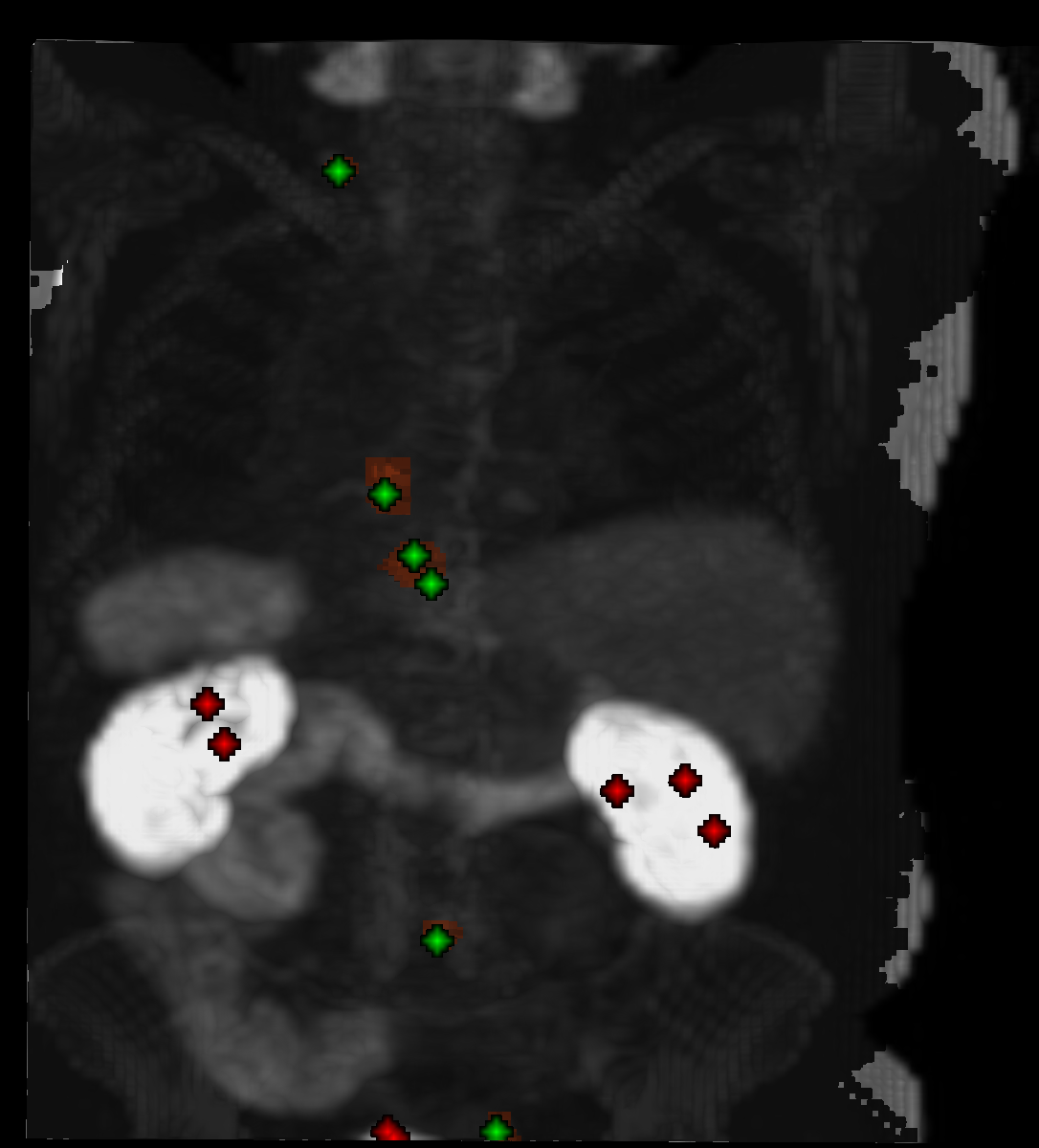}
    \end{subfigure}
    \begin{subfigure}[t]{0.32\textwidth}
        \centering\includegraphics[width=\textwidth]{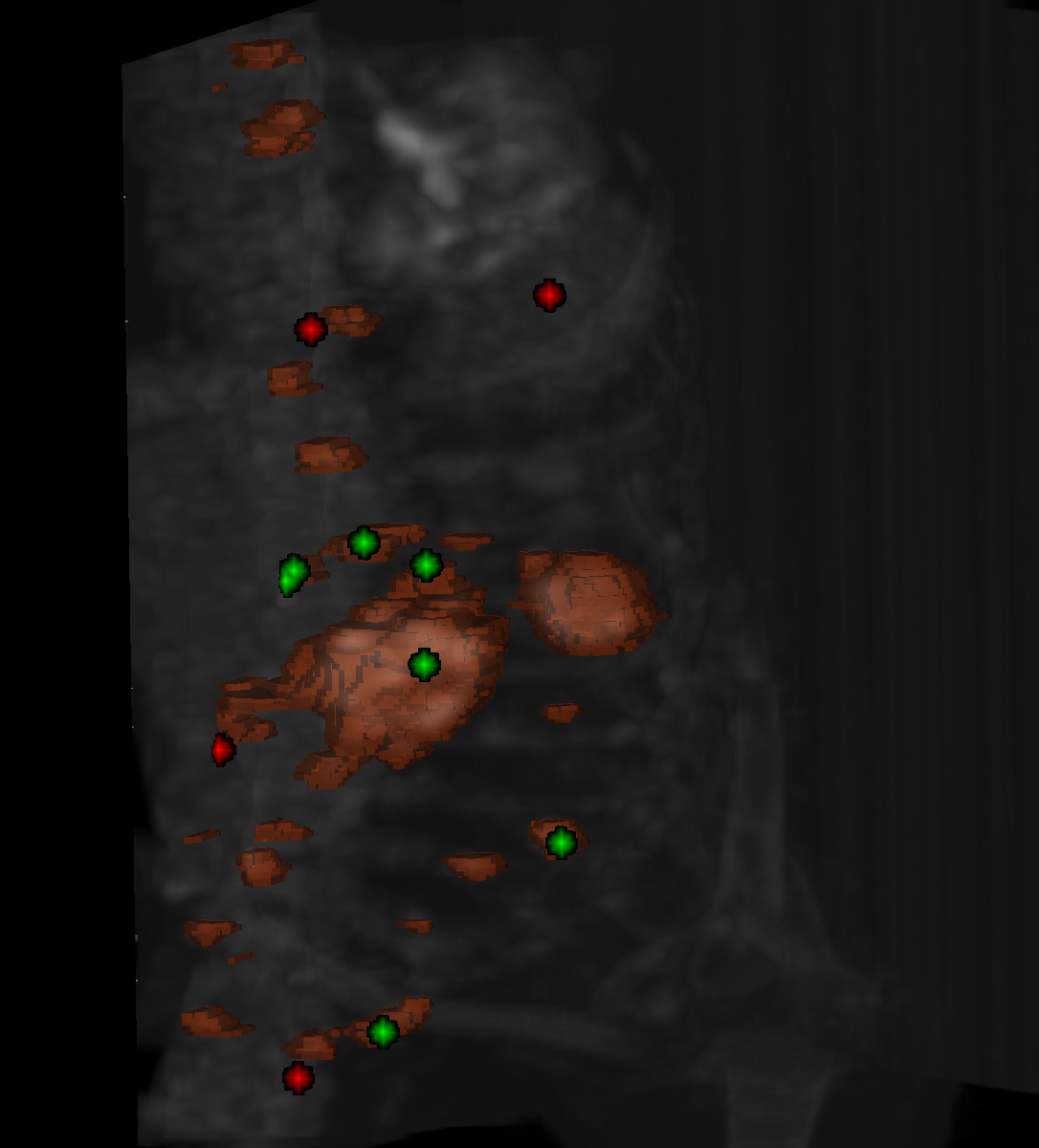}
    \end{subfigure}
    \caption{Examplary visualization of 3D image patches provided to the network during training with the respective positive (green) and negative (red) points as well as the ground truth segmentation mask shown.}
    \label{fig:organs}
\end{figure}

\subsection{Data}

\subsubsection{Training data}
In addition to the datasets provided by the autoPET/CT IV challenge~\cite{gatidis2020fdgpetct}\cite{jeblick2024psmapetct}, we incorporated several external resources to enhance model robustness and generalization. Specifically, we utilized the DEEP-PSMA dataset \cite{ferdinandus2020prognostic}, the TCGA-LUAD collection \cite{Albertina2016TCGALUAD}, and the NSCLC Radiogenomics dataset \cite{bakr2018radiogenomic}.

\subsubsection{Validation data}
For validation, we exclusively relied on the official challenge dataset and adhered to the provided splits. This ensured comparability across models and allowed us to optimize performance for the test distribution.

\begin{table}[t]
\begin{center}
    \caption{Results from five-fold cross-validation. Metrics are calculated according to the official evaluation implementation and split. Note that the AUC is computed only using 0, 3, 7, and 10 clicks.}
    \label{tab:results}
        \setlength{\tabcolsep}{4pt} 
        \begin{tabular}{l|cc|cc|cc}
            & \multicolumn{2}{c|}{Dice$\uparrow$} & \multicolumn{2}{c|}{FPvol$\downarrow$} & \multicolumn{2}{c}{FNvol$\downarrow$}\\
        Setting & AUC & Last & AUC & Last & AUC & Last  \\ \hline
        autoPET III model & - & 68.33 & - & 8.93 & - & 10.15 \\
        \hline
        + Gaussian Kernel $\sigma=3$ & 2.06 & 68.70 & 26.74& 8.92 & 18.31 &	6,14 \\
        + Gaussian Kernel $\sigma=2$ & 2.10 & 70.78 & 26.31 & 8.76 & 18.60 & 6.15 \\
        + Gaussian Kernel $\sigma=1$ & 2.12 & 71.89 & 26.34 & 8.75 & 17.83 & 5.68 \\
        + Gaussian Kernel $\sigma=0.75$ & 2.14 & 72.37 & 26.45 & 8.78 & 17.70 & 5.56 \\
        + Gaussian Kernel $\sigma=0.5$ & 2.15 & 72.94 & 25.73 & 8.55 & 16.86 & 5.23 \\
        + Gaussian Kernel $\sigma=0.25$& 2.18 & 74.53 & 25.33 & 8.39 & 15.98 & 4.60 \\
        + Gaussian Kernel $\sigma=0.1$ & 2.18 & 74.59 & 26.05 & 8.64 & 16.11 & 4.75 \\
        \hline
        + EDT Size $4$ & 2.21 & 75.75 & 26.48 & 8.73 & 13.67 & 3.66 \\
        + EDT Size $2$ & 2.22 & 76.09 & 26.31 & 8.62 & 14.46 & 3.86 \\
        + EDT Size $3$ & 2.21 & 75.89 & 26.85 & 8.81 & 15.01 & 4.06 \\
        + EDT Size $2$, custom points & 2.21 & 76.19 & \textbf{24.53} & \textbf{8.05} & 15.11 & 4.19 \\
        \hline
        + EDT Size $2$, more data & 2.21 & 76.00 & 24.85 & 8.19 & 15.45 & 4.15 \\
        + EDT Size $2$, more data, custom points & \textbf{2.22} & \textbf{76.35} & 26.03 & 8.50 & \textbf{14.17} & \textbf{3.76} \\
        \end{tabular}
    \end{center}
\end{table}

\section{Results}

Table~\ref{tab:results} reports the results of our five-fold cross-validation experiments, evaluating Dice, false positive volume (FPvol), and false negative volume (FNvol) metrics. The table compares the baseline autoPET III model with different spatial prompt representations added, including Gaussian kernels with varying standard deviations and Euclidean Distance Transform (EDT) representations of different sizes. All metrics are computed according to the official evaluation protocol, with AUC calculated using 0, 3, 7, and 10 clicks.\\

\noindent The results demonstrate that EDT-based representations consistently outperform Gaussian kernels across all metrics, particularly for Dice and FPvol. Among the EDT variants, the model with Size 2 and custom points achieves strong performance, reducing FPvol and FNvol compared to other configurations. Incorporating additional training data further improves Dice and FNvol, while the custom point sampling strategy provides better generalization to realistic click distributions.

\subsection{Test Set Submission}

For the final challenge submission, we ensemble all 5 folds of two models: (i) the EDT Size 2 model with custom points (including 20\% of points sampled using our custom strategy) and (ii) the same model trained with additional data. This ensemble leverages the complementary strengths of both models, combining the robust performance of the base EDT Size 2 model with the broader generalization afforded by additional training data. No post-processing was applied. Overall, these two approaches achieved the best performance on the cross-validation.

\section{Conclusion}

We extended the autoPET III nnU-Net model pipeline to the interactive setting of autoPET/CT IV, showing that Euclidean Distance Transform (EDT)–based click encodings outperform Gaussian kernels. Online simulation of user interactions and a custom point sampling strategy improved robustness to realistic clicks. Our final ensemble with and without additional data, achieved the best cross-validation performance. These results demonstrate the effectiveness of promptable models for human-in-the-loop PET/CT segmentation and support their use in efficient, user-guided workflows.



%
%
%
%
\newpage
\bibliographystyle{splncs04}
\bibliography{bib}

\begin{table}[ht]
\caption{Algorithm details}\label{tab1}

\begin{tabular}{P{0.2\textwidth}P{0.2\textwidth}P{0.2\textwidth}P{0.2\textwidth}P{0.2\textwidth}} 
\toprule
\textbf{Team name} & \textbf{algorithm name} (as submitted on grand-challenge) & \textbf{data pre-processing} & \textbf{data post-processing} & \textbf{training data augmentation} \\
\midrule
LesionLocator & LesionLocator &  normalization \& resampling & - & nnUNet augmentations, misalignment augmentations \\
\bottomrule
\end{tabular}

\vspace{2em}

\begin{tabular}{P{0.2\textwidth}P{0.2\textwidth}P{0.2\textwidth}P{0.2\textwidth}P{0.2\textwidth}} 
\toprule
\textbf{test time augmentation} & \textbf{ensembling} (e.g. cross-validation, model ensemble, ...) & \textbf{standardized framework?} (e.g. nnUNet, MONAI, ...)  & \textbf{network architecture} (e.g. UNet (3D)) & \textbf{loss} \\
\midrule
- & 5-folds of two best performing models & nnUNet (3D) & UNet (3D) & DSC + CE \\
& & & & \\
\bottomrule
\end{tabular}

\vspace{2em}

\begin{tabular}{P{0.2\textwidth}P{0.2\textwidth}P{0.2\textwidth}P{0.2\textwidth}P{0.2\textwidth}} 
\toprule
\textbf{training data} & \textbf{data/model dimensionality and size} & \textbf{use of pre-trained models} & \textbf{GPU hardware for training}\\
\midrule
autoPET III, DEEP-PSMA, TCGA-LUAD, NSCLC-Radiogenomics & 3D: 192x192x192 & yes, MultiTalent pretraining & Nvidia A100 \\
& & & & \\
\bottomrule
\end{tabular}
\end{table}
\end{document}